\documentclass[conference,a4paper]{IEEEtran}

\usepackage[utf8]{inputenc} 
\usepackage[T1]{fontenc}
\usepackage{url}              
\usepackage{cite}             
\usepackage[cmex10]{amsmath}  
\usepackage{mleftright}       
\mleftright                   
\usepackage{graphicx}         
\usepackage{booktabs}         

\usepackage{ISIT23_ours}

\IEEEoverridecommandlockouts
\begin{document}

\title{On the Validation of Gibbs Algorithms: Training Datasets, Test Datasets and their Aggregation}

 \author{%
   \IEEEauthorblockN{Samir M. Perlaza\IEEEauthorrefmark{1}\IEEEauthorrefmark{2}\IEEEauthorrefmark{3},
                     I\~{n}aki Esnaola\IEEEauthorrefmark{2}\IEEEauthorrefmark{4},
                     Gaetan Bisson\IEEEauthorrefmark{3},
                     and H. Vincent Poor\IEEEauthorrefmark{2}}
   \IEEEauthorblockA{\IEEEauthorrefmark{1}%
                     INRIA, Centre Inria d'Université Côte d'Azur,
                     Sophia Antipolis, France.}
   \IEEEauthorblockA{\IEEEauthorrefmark{2}%
                     ECE Dept. Princeton University, 
                     Princeton, 08544 NJ, USA.}
   \IEEEauthorblockA{\IEEEauthorrefmark{3}%
                     GAATI, Université de la Polynésie Française,
                     Faaa, French Polynesia.}
   \IEEEauthorblockA{\IEEEauthorrefmark{4}%
                     ACSE Dept., University of Sheffield, 
                     Sheffield, United Kingdom.  \thanks{This work is supported by the Inria Exploratory Action -- Information and Decision Making (AEx IDEM) and in part by a grant from the C3.ai Digital Transformation Institute.}}

  }

\maketitle

\begin{abstract}
The dependence on training data of the Gibbs algorithm (GA) is analytically characterized.  By adopting the expected empirical risk as the performance metric, the sensitivity of the GA is obtained in closed form. In this case, sensitivity is the performance difference with respect to an arbitrary alternative algorithm. This description enables the development of explicit expressions involving the training errors and test errors of GAs trained with different datasets. Using these tools, dataset aggregation is studied and different figures of merit to evaluate the generalization capabilities of GAs are introduced. For particular sizes of such datasets and parameters of the GAs, a connection between Jeffrey’s divergence, training and test errors is established.
%
\end{abstract}

\section{Introduction}
The Gibbs algorithm (GA) randomly selects a model by sampling the Gibbs probability measure, which is the unique solution to the empirical risk minimization (ERM) problem with relative entropy regularization (ERM-RER) \cite{InriaRR9454}. 
The input of the GA is twofold. It requires a number of labeled patterns (datasets);  and  a prior on the set of models in the form of a~$\sigma$-measure, e.g., the Lebesgue measure, the counting measure, or a probability measure. 
One of the main features of the GA is that it does not require an assumption on the statistical properties of the datasets \cite{catoni2004statistical, zdeborova2016statistical, alquier2016properties}. 
Nonetheless, the generalization capabilities of the Gibbs algorithm are often characterized by the generalization error, for which statistical assumptions on the datasets must be considered, e.g., training, and unseen datasets are identically distributed. When the prior on the set of models is a probability measure, a closed-form expression for the generalization error is presented in \cite{Aminiam2021Exact}, while upper bounds have been derived  in \cite{zhang2006ep, zhang2006information, jiao2017dependence, xu2017information, wang2019information, issa2019strengthened, russo2019much, bu2020tightening, asadi2018chaining, lopez2018generalization,  asadi2020chaining, hafez2020conditioning, haghifam2020sharpened, rodriguez2021tighter,  esposito2021generalization, aminian2021jensen, aminian2022information, aminian2022tighter, shawe1997pac, mcallester2003pac, haddouche2020pacbayes, guedj2019free}, and references therein.

In a more general setting, when the prior on the set of models is a~$\sigma$-measure, the generalization capabilities of the GA have been studied in \cite{Perlaza-ISIT-2022, InriaRR9454}, and \cite{InriaRR9474}, using the sensitivity of the empirical risk to deviations from the Gibbs probability measure to another probability measure. This method does not require any statistical assumptions on the datasets and is chosen as the workhorse of the present analysis.

The main motivation of this work is to break away from the implicit assumption in existing literature that all training datasets are drawn from the same probability measure and thus, can be aggregated to improve the generalization capabilities of a given GA.  In practical settings, training data might be acquired from multiple sources that might be subject to different impairments during data acquisition, data storage and data transmission. 
For instance, consider a GA trained  upon a particular dataset and assume that a new dataset from a different source is made available. Hence,  the following questions arise concerning the generalization capabilities of such a GA: 
Would such a GA generalize over the new dataset?  
Should the new dataset be aggregated to the previous dataset to build a new GA in the aim of improving generalization? How does the GA trained upon the existing dataset compare in terms of generalization with respect to a new GA trained upon the new dataset?
The answers to such questions are far from trivial.
One of the main challenges to answer such questions stems from the fact that the probability measures generating each of those datasets are unknown and potentially different due to a variety of impairments.  

This paper introduces a closed-form expression for the difference of the expected empirical risk on a given dataset induced by a GA trained upon this dataset and the one induced by an alternative algorithm (another probability measure). This quantity was coined \emph{sensitivity of the GA algorithm} in \cite{Perlaza-ISIT-2022} and is shown to be central to tackling the questions above. 
This is in part due to the fact that it allows studying the generalization capabilities of GAs based on actual datasets, which disengages from the assumption that both training and unseen data follow the same probability distribution. More specifically, by studying the sensitivity, closed-form expressions for the difference between  training error and  test error can be obtained. These expressions lead to a clearer understanding of  the roles of the size of datasets chosen for training and testing, as well as  the parameters of the GAs. 
As a byproduct, the difference between the expected empirical risk on the aggregation of two datasets induced by two GAs trained upon the constituent  datasets is characterized. Similarly, the difference between the expected empirical risk on one of the constituent datasets induced by two GAs trained upon the aggregated dataset and the constituent dataset is also characterized.  These explicit expressions allow comparing two GAs trained upon different datasets, which is relevant under learning paradigms such as federated learning \cite{mcmahan2017communication}. 

\section{Problem Formulation}

Let~$\set{M}$,~$\set{X}$ and~$\set{Y}$, with~$\set{M} \subseteq \reals^{d}$ and~$d \in \ints$, be sets of \emph{models}, \emph{patterns}, and \emph{labels}, respectively.  
A pair~$(x,y) \in \mathcal{X} \times \mathcal{Y}$ is referred to as a \emph{labeled pattern} or as a \emph{data point}.
Given~$n$ data points, with~$n \in \ints$,  denoted by~$\left(x_1, y_1 \right)$, $\left( x_2, y_2\right)$, $\ldots$, $\left( x_n, y_n \right)$, a dataset is represented by the tuple~$\big(\left(x_1, y_1 \right)$, $\left(x_2, y_2 \right)$, $\ldots$, $\left(x_n, y_n \right)\big)$  $\in \left( \mathcal{X} \times \mathcal{Y} \right)^n$.

Let the function~$f: \set{M} \times \mathcal{X} \rightarrow \mathcal{Y}$ be such that the label~$y$ assigned to the pattern $x$ according to the model $\vect{\theta} \in \set{M}$ is 
\begin{equation}\label{EqTheModel}
    y = f(\vect{\theta}, x).
\end{equation}
Let also the function 
\begin{equation}\label{EqEll}
\ell: \set{Y} \times \set{Y} \rightarrow [0, +\infty]
\end{equation} 
be such that given a data point~$(x, y) \in \set{X} \times \set{Y}$, the  risk induced by a model~$\vect{\theta} \in \set{M}$ is ~$\ell\left( f(\vect{\theta}, x), y \right)$.  
In the following, the risk function~$\ell$ is assumed to be nonnegative and  for all~$y \in \set{Y}$, ~$\ell\left( y , y\right) = 0$.

Given a dataset 
\begin{equation}\label{EqTheDataSet}
\vect{z} = \big(\left(x_1, y_1 \right), \left(x_2, y_2 \right), \ldots, \left(x_n, y_n \right)\big)  \in \left( \set{X} \times \set{Y} \right)^n,
\end{equation}  
the \emph{empirical risk} induced by the model~$\vect{\theta}$, with respect to the dataset $\vect{z}$ in \eqref{EqTheDataSet}, is determined by the  function~$\mathsf{L}_{\vect{z}}: \set{M} \rightarrow [0, +\infty ]$, which satisfies  
\begin{IEEEeqnarray}{rCl}
\label{EqLxy}
\mathsf{L}_{\vect{z}} \left(\vect{\theta} \right)  & = & 
\frac{1}{n}\sum_{i=1}^{n}  \ell\left( f(\vect{\theta}, x_i), y_i\right).
\end{IEEEeqnarray}
Using this notation, the ERM problem consists of the following optimization problem:
\begin{equation}\label{EqOriginalOP}
\min_{\vect{\theta} \in \set{M}} \mathsf{L}_{\vect{z}} \left(\vect{\theta} \right).
\end{equation}
Let the set of solutions to the ERM problem in \eqref{EqOriginalOP} be denoted by
$\set{T}\left( \vect{z} \right) \triangleq \arg\min_{\vect{\theta} \in \set{M}}    \mathsf{L}_{\vect{z}} \left(\vect{\theta} \right)$.
%
Note that if the set $\set{M}$ is finite, the ERM problem in \eqref{EqOriginalOP} always possesses a solution, and thus, $\abs{\set{T}\left( \vect{z} \right)} > 0$. Nonetheless, in general, the ERM problem might not necessarily possess a solution. Hence, for some cases, it might be observed that $\abs{\set{T}\left( \vect{z} \right) } = 0$.

\subsection{Notation}

The \emph{relative entropy} is defined below as the extension to~$\sigma$-finite measures of the relative entropy usually defined for probability measures.

%
\begin{definition}[Relative Entropy]\label{DefRelEntropy}
Given two~$\sigma$-finite measures~$P$ and~$Q$ on the same measurable space, such that~$Q$ is absolutely continuous with respect to~$P$,  the relative entropy of~$Q$ with respect to~$P$ is
\begin{equation}
\KL{Q}{P} \triangleq \int \frac{\mathrm{d}Q}{\mathrm{d}P}(x)  \log\left( \frac{\mathrm{d}Q}{\mathrm{d}P}(x)\right)  \mathrm{d}P(x),
\end{equation}
where the function~$\frac{\mathrm{d}Q}{\mathrm{d}P}$ is the Radon-Nikodym derivative of~$Q$ with respect to~$P$.
\end{definition}
Given a measurable space~$\left( \Omega, \mathscr{F} \right)$, the set of all~$\sigma$-finite measures on~$\left( \Omega, \mathscr{F} \right)$ is denoted by~$\triangle\left( \Omega, \mathscr{F} \right)$.  Given a~$\sigma$-measure~$Q \in \triangle\left( \Omega, \mathscr{F} \right)$, the subset of~$\triangle\left( \Omega, \mathscr{F} \right)$ including all~$\sigma$-finite measures absolutely continuous with~$Q$ is denoted by~$\triangle_{Q}\left( \Omega, \mathscr{F} \right)$. Given a subset~$\set{A}$ of~$\reals^{d}$, the Borel~$\sigma$-field on~$\set{A}$ is denoted by~$\BorSigma{\set{A}}$.
 
\subsection{The ERM-RER Problem}
%
The \emph{expected empirical risk} is defined as follows.
\begin{definition}[Expected Empirical Risk]\label{DefEmpiricalRisk}
Let $P$ be a probability measure in $\Delta\Bormeaspace{\set{M}}$. The expected empirical risk with respect to the dataset~$\vect{z}$ in~\eqref{EqTheDataSet} induced by the measure $P$ is 
\begin{equation}
\label{EqRxy}
\mathsf{R}_{\vect{z}}\left( P  \right) = \int \mathsf{L}_{ \vect{z} } \left(\vect{\theta} \right)  \mathrm{d} P(\vect{\theta}),
\end{equation}
where the function~$\mathsf{L}_{\vect{z}}$ is in~\eqref{EqLxy}. 
\end{definition}

The following lemma follows immediately from the properties of the Lebesgue integral. 
\begin{lemma}\label{LemmaLinearity}
Given a dataset~$\vect{z} \in \left( \set{X} \times \set{Y} \right)^n$ and two probability measures~$P_1$ and~$P_2$ over the measurable space~$\Bormeaspace{\set{M}}$, for all~$\alpha \in [0,1]$, the function~$\mathsf{R}_{\vect{z}}$ in~\eqref{EqRxy} satisfies
\begin{IEEEeqnarray}{rcl}
\mathsf{R}_{\vect{z}}\left( \alpha P_1 + (1- \alpha)P_2\right) & = & \alpha \mathsf{R}_{\vect{z}}\left( P_1 \right) + (1- \alpha) \mathsf{R}_{\vect{z}}\left( P_2\right).
\end{IEEEeqnarray}
\end{lemma}
The ERM-RER problem is parametrized by a~$\sigma$-finite measure on~$\Bormeaspace{\set{M}}$ and a positive real, which are referred to as the \emph{reference measure} and the \emph{regularization factor}, respectively.
Let~$Q$ be a~$\sigma$-finite measure on~$\Bormeaspace{\set{M}}$ and let~$\lambda > 0$ be a positive real. The ERM-RER problem, with parameters~$Q$ and~$\lambda$, consists in the following optimization problem:
\begin{equation}\label{EqERMRER}
    \min_{P \in \triangle_{Q}\Bormeaspace{\set{M}}}  \mathsf{R}_{\vect{z}} \left( P \right)  + \lambda D\left( P \|Q\right),
\end{equation}
where the dataset~$\vect{z}$ is in~\eqref{EqTheDataSet}; and the function~$\mathsf{R}_{\vect{z}}$ is defined in~\eqref{EqRxy}.
For the ease of presentation, the parameters of the ERM-RER problem in~\eqref{EqERMRER} are chosen such that 
\begin{equation}\label{EqZeroInfinity}
 Q\left( \left\lbrace \vect{\theta} \in \set{M}:  \mathsf{L}_{\vect{z}}\left( \vect{\theta}\right) = +\infty \right\rbrace  \right) =0.
 \end{equation} 
 The case in which the regularization is $\KL{Q}{P}$ (instead of $\KL{P}{Q}$) in \eqref{EqERMRER} is left out of the scope of this work. The interested reader is referred to~\cite{Perlaza-ISIT2023a}.

\subsection{The Solution to the ERM-RER Problem}

The solution to the ERM-RER problem in~\eqref{EqERMRER} is presented by the following lemma.
\begin{lemma}[Theorem~$2.1$ in \cite{Perlaza-ISIT-2022}]\label{TheoremOptimalModel}
Given a~$\sigma$-finite measure~$Q$ and  a dataset~$\vect{z} \in \left( \set{X} \times \set{Y} \right)^n$, 
let the function~$K_{Q,\vect{z}}: \reals \rightarrow \reals \cup \lbrace +\infty\rbrace$ be such that for all~$t \in \reals$,
\begin{IEEEeqnarray}{rcl}
\label{EqK}
K_{Q,\vect{z}}\left(t \right) & = &  \log\left( \int \exp\left( t \; \mathsf{L}_{\vect{z}}\left(\vect{\theta}\right)  \right) \mathrm{d}Q(\vect{\theta}) \right),
\end{IEEEeqnarray} 
where the function~$\mathsf{L}_{\vect{z}}$ is defined in~\eqref{EqLxy}.
Let also the set~$\set{K}_{Q,\vect{z}} \subset (0, +\infty)$ be 
\begin{IEEEeqnarray}{rcl}
\label{EqSetKxy}
\set{K}_{Q,\vect{z}} & \triangleq &\left\lbrace s > 0: \; K_{Q,\vect{z}}\left(-\frac{1}{s} \right)  < +\infty \right\rbrace.
\end{IEEEeqnarray}
Then, for all~$\lambda \in \set{K}_{Q,\vect{z}}$, the solution to the ERM-RER problem in~\eqref{EqERMRER} is a unique measure on~$\Bormeaspace{\set{M}}$, denoted by 
$P^{\left(Q, \lambda\right)}_{\vect{\Theta}| \vect{Z} = \vect{z}}$, whose Radon-Nikodym derivative with respect to~$Q$ satisfies that for all~$\vect{\theta} \in \supp Q$,
\begin{IEEEeqnarray}{rcl}\label{EqGenpdf}
\frac{\mathrm{d}P^{\left(Q, \lambda\right)}_{\vect{\Theta}| \vect{Z} = \vect{z}}}{\mathrm{d}Q} \left( \vect{\theta} \right) 
  & =& \exp\left( - K_{Q,\vect{z}}\left(- \frac{1}{\lambda} \right) - \frac{1}{\lambda} \mathsf{L}_{\vect{z}}\left( \vect{\theta}\right)\right).
\end{IEEEeqnarray}
\end{lemma}
Among the numerous properties of the solution to the ERM-RER problem in~\eqref{EqERMRER}, the following property is particularly useful in the remainder of this work.  
\begin{lemma}\label{LemmaAgadir}
Given a~$\sigma$-finite measure~$Q$ over the measurable space~$\Bormeaspace{\set{M}}$, and given a dataset~$\vect{z} \in \left( \set{X} \times \set{Y} \right)^n$, for all~$\lambda \in \set{K}_{Q,\vect{z}}$, with~$\set{K}_{Q,\vect{z}}$ in~\eqref{EqSetKxy}, the following holds:
\begin{IEEEeqnarray}{rcl}
\label{EqAgadirEarlyMorningA}
\mathsf{R}_{\vect{z}}\left( P^{\left(Q, \lambda\right)}_{\vect{\Theta}| \vect{Z} = \vect{z}} \right)  + \lambda \KL{P^{\left(Q, \lambda\right)}_{\vect{\Theta}| \vect{Z} = \vect{z}}}{Q} & = & - \lambda K_{Q,\vect{z}}\left(- \frac{1}{\lambda} \right),
\end{IEEEeqnarray}
where
the function~$\mathsf{R}_{\vect{z}}$ is defined in~\eqref{EqRxy};  
the function~$K_{Q,\vect{z}}$ is defined in~\eqref{EqK}; and 
the probability measure~$P^{\left(Q, \lambda\right)}_{\vect{\Theta}| \vect{Z} = \vect{z}}$ is the solution to the ERM-RER problem in~\eqref{EqERMRER}.
\end{lemma}
\begin{IEEEproof}
The proof is presented in~\cite{InriaRR9474}.
\end{IEEEproof} 
 \section{Sensitivity of the ERM-RER Solution}
The sensitivity of the expected empirical risk~$\mathsf{R}_{\vect{z}}$ to deviations from the probability measure~$P^{\left(Q, \lambda\right)}_{\vect{\Theta}| \vect{Z} = \vect{z}}$ towards an alternative probability measure~$P$ is defined as follows. 
\begin{definition}[Sensitivity \cite{Perlaza-ISIT-2022}]\label{DefSensitivity}
Given a~$\sigma$-finite measure~$Q$ and a positive real~$\lambda > 0$, let~$\mathsf{S}_{Q, \lambda}: \left( \set{X} \times \set{Y} \right)^n \times \triangle_{Q}\Bormeaspace{\set{M}}\rightarrow \left( - \infty, +\infty \right]$ be a function such that \begin{IEEEeqnarray}{l}
\label{EqDefSensitivity}
\mathsf{S}_{Q, \lambda}\left( \vect{z}, P \right)  =
\left\lbrace
\begin{array}{cl}
\mathsf{R}_{\vect{z}}\left( P \right)  - \mathsf{R}_{\vect{z}}\left( P^{\left(Q, \lambda\right)}_{\vect{\Theta}| \vect{Z} = \vect{z}} \right) & \text{ if } \lambda \in \set{K}_{Q,\vect{z}}\\
+\infty & \text{ otherwise,}
\end{array}
\right. \quad
\end{IEEEeqnarray}
where the function~$\mathsf{R}_{\vect{z}}$ is defined in~\eqref{EqRxy} and the measure~$P^{\left(Q, \lambda\right)}_{\vect{\Theta}| \vect{Z} = \vect{z}}$ is the solution to the ERM-RER  problem in~\eqref{EqERMRER}.
The sensitivity of the expected empirical risk~$\mathsf{R}_{\vect{z}}$ when the measure changes from~$P^{\left(Q, \lambda\right)}_{\vect{\Theta}| \vect{Z} = \vect{z}}$ to~$P$ is~$\mathsf{S}_{Q, \lambda}\left( \vect{z}, P\right)$.
\end{definition}
The following theorem introduces an exact expression for the sensitivity in Definition~\ref{DefSensitivity}.
\begin{theorem}\label{TheoremSensitivityEqual}
Given a~$\sigma$-finite measure~$Q$ over the measurable space~$\Bormeaspace{\set{M}}$ and a probability measure~$P \in \triangle_{Q}\Bormeaspace{\set{M}}$, it holds that 
 for all datasets~$\vect{z} \in \left( \set{X} \times \set{Y} \right)^n$ and for all~$\lambda \in \set{K}_{Q,\vect{z}}$, with~$\set{K}_{Q,\vect{z}}$ in~\eqref{EqSetKxy}, 
\begin{IEEEeqnarray}{rcl}
\nonumber
\mathsf{S}_{Q, \lambda}\left( \vect{z}, P \right)& = &  \lambda\Big( \KL{P^{\left(Q, \lambda\right)}_{\vect{\Theta}| \vect{Z} = \vect{z}}}{Q} + \KL{P}{P^{\left(Q, \lambda\right)}_{\vect{\Theta}| \vect{Z} = \vect{z}}} - \KL{P}{Q} \Big),
\end{IEEEeqnarray}
where  the probability measure~$P^{\left(Q, \lambda\right)}_{\vect{\Theta}| \vect{Z} = \vect{z}}$ is the solution to the ERM-RER problem in~\eqref{EqERMRER}.
\end{theorem}
\begin{IEEEproof}
The proof uses the fact that, under the assumption in~\eqref{EqZeroInfinity}, the probability measure~$P^{\left(Q, \lambda\right)}_{\vect{\Theta}| \vect{Z} = \vect{z}}$ in~\eqref{EqGenpdf} is mutually absolutely continuous with respect to the~$\sigma$-finite measure~$Q$; see for instance \cite{InriaRR9454}. Hence, the probability measure~$P$ is absolutely continuous with respect to~$P^{\left(Q, \lambda\right)}_{\vect{\Theta}| \vect{Z} = \vect{z}}$, as a consequence of the assumption that~$P$ is absolutely continuous with respect to~$Q$.

The proof follows by noticing that for all~$\vect{\theta} \in \set{M}$, 
\begin{IEEEeqnarray}{rcl}
&&\log\left(\frac{\mathrm{d}P}{\mathrm{d}P^{\left(Q, \lambda\right)}_{\vect{\Theta}| \vect{Z} = \vect{z}}} \left( \vect{\theta} \right)\right) =  \log\left(\frac{\mathrm{d}Q}{\mathrm{d}P^{\left(Q, \lambda\right)}_{\vect{\Theta}| \vect{Z} = \vect{z}}} \left( \vect{\theta} \right) \frac{\mathrm{d}P}{\mathrm{d}Q} \left( \vect{\theta} \right)\right) \qquad \\
& = & - \log\left(\frac{\mathrm{d}P^{\left(Q, \lambda\right)}_{\vect{\Theta}| \vect{Z} = \vect{z}}}{\mathrm{d}Q} \left( \vect{\theta} \right) \right) + \log\left( \frac{\mathrm{d}P}{\mathrm{d}Q} \left( \vect{\theta} \right)\right) \\
\label{EqMorningSouad}
& = &  K_{Q,\vect{z}}\left(- \frac{1}{\lambda} \right) + \frac{1}{\lambda} \mathsf{L}_{\vect{z}}\left( \vect{\theta}\right)+ \log\left( \frac{\mathrm{d}P}{\mathrm{d}Q} \left( \vect{\theta} \right)\right),
\end{IEEEeqnarray}
where the functions~$\mathsf{L}_{\vect{z}}$ and~$K_{Q,\vect{z}}$ are defined in~\eqref{EqLxy} and in~\eqref{EqK}, respectively; and 
the equality in~\eqref{EqMorningSouad} follows from Lemma~\ref{TheoremOptimalModel}.
Hence, the relative entropy~$\KL{P}{P^{\left(Q, \lambda\right)}_{\vect{\Theta}| \vect{Z} = \vect{z}}}$ satisfies
\begin{IEEEeqnarray}{rcl}
&&\KL{P}{P^{\left(Q, \lambda\right)}_{\vect{\Theta}| \vect{Z} = \vect{z}}} = \int  \log\left(\frac{\mathrm{d}P}{\mathrm{d}P^{\left(Q, \lambda\right)}_{\vect{\Theta}| \vect{Z}= \vect{z}}} \left( \vect{\theta} \right)\right) \mathrm{d}P \left( \vect{\theta} \right) \nonumber\\
\label{EqTheProofBroadCast0}
&=&   K_{Q,\vect{z}}\left(- \frac{1}{\lambda} \right) + \int \left( \frac{1}{\lambda} \mathsf{L}_{\vect{z}}\left( \vect{\theta}\right) +  \log\left( \frac{\mathrm{d}P}{\mathrm{d}Q} \left( \vect{\theta} \right)\right) \right)\mathrm{d}P \left( \vect{\theta} \right) \quad\\
\label{EqTheProofBroadCast1}
& = &  K_{Q,\vect{z}}\left(- \frac{1}{\lambda} \right) + \frac{1}{\lambda}\mathsf{R}_{\vect{z}}\left( P \right) + \KL{P}{Q}\\
\nonumber
& = & - \KL{P^{\left(Q, \lambda\right)}_{\vect{\Theta}| \vect{Z} = \vect{z}}}{Q} + \frac{1}{\lambda} \left( \mathsf{R}_{\vect{z}}\left( P \right) - \mathsf{R}_{\vect{z}}\left( P^{\left(Q, \lambda\right)}_{\vect{\Theta}| \vect{Z} = \vect{z}} \right)  \right) \\
& & + \KL{P}{Q}, \label{EqTheProofBroadCast2}
\end{IEEEeqnarray}
where the function~$\mathsf{R}_{\vect{z}}$  is defined in~\eqref{EqRxy}, 
the equality in~\eqref{EqTheProofBroadCast0} follows from~\eqref{EqMorningSouad}, and 
the equality in~\eqref{EqTheProofBroadCast2} follows from Lemma~\ref{LemmaAgadir}.
Finally, the proof is completed by re-arranging the terms in~\eqref{EqTheProofBroadCast2}.
\end{IEEEproof}

\section{Validation of Gibbs Algorithms}

Consider the dataset~$\vect{z}_{0} \in \left( \set{X} \times \set{Y} \right)^{n_{0}}$ that aggregates dataset~$\vect{z}_{1}\in \left( \set{X} \times \set{Y} \right)^{n_{1}}$ and dataset~$\vect{z}_{2}\in \left( \set{X} \times \set{Y} \right)^{n_{2}}$ as constituents. That is,~$
\vect{z}_0 = \left( \vect{z}_1, \vect{z}_2 \right)$,  with~$n_{0} = n_{1} + n_{2}$. 
Datasets~$\vect{z}_{1}$ and~$\vect{z}_{2}$ are referred to as \emph{constituent datasets}, whereas, the dataset~$\vect{z}_{0}$ is referred to as the \emph{aggregated dataset}.
For all~$i \in \lbrace 0,1,2 \rbrace$, the empirical risk function in~\eqref{EqLxy} and the expected empirical risk function  in~\eqref{EqRxy} over dataset~$\vect{z}_{i}$ are denoted by~$\mathsf{L}_{\vect{z}_{i}}$ and~$\mathsf{R}_{\vect{z}_{i}}$, respectively. Such functions exhibit the following property. 
\begin{lemma}\label{LemmaSumLxy}
The empirical risk functions~$\mathsf{L}_{\vect{z}_{0}}$,~$\mathsf{L}_{\vect{z}_{1}}$, and~$\mathsf{L}_{\vect{z}_{2}}$, defined in~\eqref{EqLxy} satisfy for all~$\vect{\theta} \in \set{M}$,
\begin{IEEEeqnarray}{rcl}
\label{EqYTuYTu}
\mathsf{L}_{\vect{z}_{0}} \left( \vect{\theta} \right) & = & \frac{n_1}{n_0}\mathsf{L}_{\vect{z}_{1}}\left( \vect{\theta} \right) + \frac{n_2}{n_0}\mathsf{L}_{\vect{z}_{2}}\left( \vect{\theta} \right).
\end{IEEEeqnarray}
Moreover, the expected empirical risk functions~$\mathsf{R}_{\vect{z}_{0}}$,~$\mathsf{R}_{\vect{z}_{1}}$, and~$\mathsf{R}_{\vect{z}_{2}}$, defined in~\eqref{EqRxy}, satisfy for all~$\sigma$-finite measures~$P \in \triangle\Bormeaspace{\set{M}}$, 
\begin{IEEEeqnarray}{rcl}
\label{EqYTuYTuConR}
\mathsf{R}_{\vect{z}_{0}} \left( P \right) & = & \frac{n_1}{n_0}\mathsf{R}_{\vect{z}_{1}}\left( P \right) + \frac{n_2}{n_0}\mathsf{R}_{\vect{z}_{2}}\left( P \right).
\end{IEEEeqnarray}
\end{lemma}
\begin{IEEEproof}
The proof is presented in~\cite{InriaRR9474}.
\end{IEEEproof} 
For all~$i \in \lbrace 0,1,2 \rbrace$, let~$Q_{i} \in \triangle\Bormeaspace{\set{M}}$ and~$\lambda_{i} \in \set{K}_{Q_{i},\vect{z}_{i}}$, with~$\set{K}_{Q_{i},\vect{z}_{i}}$ in~\eqref{EqSetKxy}, be the~$\sigma$-finite measure acting as the reference measure  and regularization factor for the learning task with dataset~$i$, respectively.
Each dataset induces a different ERM-RER problem formulation of the form
\begin{equation}\label{EqERMRERi}
    \min_{P \in \triangle_{Q_{i}} \Bormeaspace{\set{M}}}  \mathsf{R}_{\vect{z}_i} \left(P \right)  + \lambda_i D\left( P\|Q_{i}\right),
\end{equation}
where~$\mathsf{R}_{\vect{z}_i}$ is the expected empirical risk defined in~\eqref{EqRxy}.
For all~$i \in \lbrace 0,1,2 \rbrace$, the solution to the ERM-RER problem in~\eqref{EqERMRERi} is the probability measure denoted by~$P^{\left(Q_{i}, \lambda_i\right)}_{\vect{\Theta}| \vect{Z} = \vect{z}_{i}}$. 
In particular, from Lemma~\ref{TheoremOptimalModel}, it holds that the probability measure~$P^{\left(Q_{i}, \lambda_i\right)}_{\vect{\Theta}| \vect{Z} = \vect{z}_{i}}$ satisfies for all~$\vect{\theta} \in \supp Q_{i}$, 
\begin{IEEEeqnarray}{rcl}\label{EqGenpdfi}
\frac{\mathrm{d}P^{\left(Q_{i}, \lambda_{i}\right)}_{\vect{\Theta}| \vect{Z} = \vect{z}_{i}}}{\mathrm{d}Q_{i}} \left( \vect{\theta} \right) 
  & =& \exp\left( - K_{Q_{i},\vect{z}_{i}}\left(- \frac{1}{\lambda_{i}} \right) - \frac{1}{\lambda_{i}} \mathsf{L}_{\vect{z}_{i}}\left( \vect{\theta}\right)\right).
\end{IEEEeqnarray}

For all~$i \in \lbrace 0,1,2\rbrace$, the probability measure~$P^{\left(Q_{i}, \lambda_{i}\right)}_{\vect{\Theta}| \vect{Z} = \vect{z}_{i}}$ in~\eqref{EqGenpdfi} represents a GA trained upon the dataset~$\vect{z}_{i}$ with parameters~$\left(Q_{i}, \lambda_{i}\right)$. In the following, such an algorithm is denoted by~$\mathsf{GA}_i$ and the dataset~$\vect{z}_{i}$ is often referred to as the \emph{training dataset} of~$\mathsf{GA}_i$. The dataset~$\vect{z}_{j}$, with $j \in \lbrace 0,1,2\rbrace\setminus\lbrace i \rbrace$, which might contain datapoints that are not in~$\vect{z}_{i}$, is referred to as the \emph{test dataset} for~$\mathsf{GA}_i$.

\subsection{Gibbs Algorithms Trained on Constituent Datasets}

The expected empirical risk induced by~$\mathsf{GA}_i$ on the training dataset~$\vect{z}_{i}$  is the \emph{training expected empirical risk}, which is denoted by~$\mathsf{R}_{\vect{z}_{i}}\left( P^{\left(Q_{i}, \lambda_{i}\right)}_{\vect{\Theta}| \vect{Z} = \vect{z}_{i}} \right)$ and  often referred to as the \emph{training error} \cite{shalev2014understanding}. 
Alternatively,  the expected empirical risk induced by~$\mathsf{GA}_i$ on the test dataset~$\vect{z}_{j}$  is the \emph{test expected empirical risk}, which is denoted by~$\mathsf{R}_{\vect{z}_{j}}\left( P^{\left(Q_{i}, \lambda_{i}\right)}_{\vect{\Theta}| \vect{Z} = \vect{z}_{i}} \right)$ and  often referred to as the \emph{test error} \cite{shalev2014understanding}. 
%
%
The following theorem provides explicit expressions involving the training and test errors of~$\mathsf{GA}_1$ and~$\mathsf{GA}_2$. 

\begin{theorem}\label{TheoBlancNaive}
Assume that the~$\sigma$-finite measures~$Q_1$ and~$Q_2$ in~\eqref{EqERMRERi} are mutually absolutely continuous. Then, for all~$i \in \lbrace 1,2\rbrace$ and~$j \in \lbrace 1,2\rbrace\setminus\lbrace i \rbrace$,
\begin{IEEEeqnarray}{rcl}
\nonumber
& & \mathsf{R}_{\vect{z}_{i}}\left( P^{\left(Q_{j}, \lambda_{j}\right)}_{\vect{\Theta}| \vect{Z} = \vect{z}_{j}}\right)  -  \mathsf{R}_{\vect{z}_{i}}\left( P^{\left(Q_{i}, \lambda_{i}\right)}_{\vect{\Theta}| \vect{Z} = \vect{z}_{i}} \right) = \lambda_{i} \Big( \KL{P^{\left(Q_{i}, \lambda_{i}\right)}_{\vect{\Theta}| \vect{Z} = \vect{z}_{i}}}{Q_{i}} \\ 
\label{EqPizza018}
& &+  \KL{P^{\left(Q_j, \lambda_j\right)}_{\vect{\Theta}| \vect{Z} = \vect{z}_{j}}}{P^{\left(Q_i, \lambda_i\right)}_{\vect{\Theta}| \vect{Z} = \vect{z}_{i}}}  
 -  \KL{P^{\left(Q_{j}, \lambda_{j}\right)}_{\vect{\Theta}| \vect{Z} = \vect{z}_{j}} }{Q_i} \Big),\quad
\end{IEEEeqnarray}
where the function~$\mathsf{R}_{\vect{z}_{i}}$ is defined in~\eqref{EqRxy} and the measure~$P^{\left(Q_{i}, \lambda_{i}\right)}_{\vect{\Theta}| \vect{Z} = \vect{z}_{i}}$ satisfies~\eqref{EqGenpdfi}.
\end{theorem}
\begin{IEEEproof}
The proof is immediate from Theorem~\ref{TheoremSensitivityEqual} by noticing that for all~$i \in \lbrace 1,2 \rbrace$ and for all~$j \in \lbrace 1,2 \rbrace\setminus\lbrace i \rbrace$, the differences~$\mathsf{R}_{\vect{z}_{i}}\left( P^{\left(Q_{j}, \lambda_{j}\right)}_{\vect{\Theta}| \vect{Z} = \vect{z}_{j}}\right) - \mathsf{R}_{\vect{z}_{i}}\left( P^{\left(Q_{i}, \lambda_{i}\right)}_{\vect{\Theta}| \vect{Z} = \vect{z}_{i}}\right)$ can be written in terms of the sensitivity~$\mathsf{S}_{Q_i, \lambda_i}\left( \vect{z}_i, P^{\left(Q_{j}, \lambda_{j}\right)}_{\vect{\Theta}| \vect{Z} = \vect{z}_{j}} \right)$.
\end{IEEEproof}

A reasonable figure of merit to compare two machine learning algorithms trained upon two different training datasets is the difference between the expected empirical risk they induce upon the aggregation of their training datasets. The following theorem provides an explicit expression for this figure of merit for the case of the algorithms~$\mathsf{GA}_1$ and~$\mathsf{GA}_2$.

\begin{theorem}\label{TheoThreeYearAnniversary}
Assume that the~$\sigma$-finite measures~$Q_1$ and~$Q_2$ in~\eqref{EqERMRERi} are mutually absolutely continuous. Then, 
\begin{IEEEeqnarray}{rcl}
\nonumber
& & \mathsf{R}_{\vect{z}_{0}}\left( P^{\left(Q_{2}, \lambda_{2}\right)}_{\vect{\Theta}| \vect{Z} = \vect{z}_{2}}\right)  -  \mathsf{R}_{\vect{z}_{0}}\left( P^{\left(Q_{1}, \lambda_{1}\right)}_{\vect{\Theta}| \vect{Z} = \vect{z}_{1}} \right) = \frac{n_1}{n_0}\lambda_{1} \bigg( \KL{P^{\left(Q_{1}, \lambda_{1}\right)}_{\vect{\Theta}| \vect{Z} = \vect{z}_{1}}}{Q_{1}} \\ 
\nonumber
& &+  \KL{P^{\left(Q_2, \lambda_2\right)}_{\vect{\Theta}| \vect{Z} = \vect{z}_{2}}}{P^{\left(Q_1, \lambda_1\right)}_{\vect{\Theta}| \vect{Z} = \vect{z}_{1}}}   -  \KL{P^{\left(Q_{2}, \lambda_{2}\right)}_{\vect{\Theta}| \vect{Z} = \vect{z}_{2}} }{Q_1} \bigg) \quad\\
\nonumber
& &- \frac{n_2}{n_0}\lambda_{2} \bigg( \KL{P^{\left(Q_{2}, \lambda_{2}\right)}_{\vect{\Theta}| \vect{Z} = \vect{z}_{2}}}{Q_{2}} +  \KL{P^{\left(Q_1, \lambda_1\right)}_{\vect{\Theta}| \vect{Z} = \vect{z}_{1}}}{P^{\left(Q_2, \lambda_2\right)}_{\vect{\Theta}| \vect{Z} = \vect{z}_{2}}} \\
\label{EqThreeYearsEsnaola}
& &  -  \KL{P^{\left(Q_{1}, \lambda_{1}\right)}_{\vect{\Theta}| \vect{Z} = \vect{z}_{1}} }{Q_2} \bigg),\quad
\end{IEEEeqnarray}
where  the function~$\mathsf{R}_{\vect{z}_{0}}$ is defined in~\eqref{EqRxy} and, for all~$i \in \lbrace 1,2\rbrace$, the measure~$P^{\left(Q_{i}, \lambda_{i}\right)}_{\vect{\Theta}| \vect{Z} = \vect{z}_{i}}$ satisfies~\eqref{EqGenpdfi}.
\end{theorem}
\begin{IEEEproof}
The proof uses the following argument:
\begin{IEEEeqnarray}{rcl}
\nonumber
&& \mathsf{R}_{\vect{z}_{0}}\left( P^{\left(Q_{2}, \lambda_{2}\right)}_{\vect{\Theta}| \vect{Z} = \vect{z}_{2}} \right)   - \mathsf{R}_{\vect{z}_{0}}\left( P^{\left(Q_{1}, \lambda_{1}\right)}_{\vect{\Theta}| \vect{Z} = \vect{z}_{1}}  \right) \\
\nonumber
& = & \frac{n_1}{n_0} \mathsf{R}_{\vect{z}_{1}}\left( P^{\left(Q_{2}, \lambda_{2}\right)}_{\vect{\Theta}| \vect{Z} = \vect{z}_{2}}  \right)   + \frac{n_2}{n_0}  \mathsf{R}_{\vect{z}_{2}}\left( P^{\left(Q_{2}, \lambda_{2}\right)}_{\vect{\Theta}| \vect{Z} = \vect{z}_{2}}  \right) \\
\label{EqFriendGoneToTexas1}
& & - \left( \frac{n_1}{n_0} \mathsf{R}_{\vect{z}_{1}}\left( P^{\left(Q_{1}, \lambda_{1}\right)}_{\vect{\Theta}| \vect{Z} = \vect{z}_{1}}  \right)   + \frac{n_2}{n_0}  \mathsf{R}_{\vect{z}_{2}}\left( P^{\left(Q_{1}, \lambda_{1}\right)}_{\vect{\Theta}| \vect{Z} = \vect{z}_{1}}  \right) \right) \\
\label{EqFriendGoneToTexas2}
& = & \frac{n_1}{n_0} \mathsf{S}_{Q_1, \lambda_1}\left( \vect{z}_1, P^{\left(Q_{2}, \lambda_{2}\right)}_{\vect{\Theta}| \vect{Z} = \vect{z}_{2}} \right) - \frac{n_2}{n_0} \mathsf{S}_{Q_2, \lambda_2}\left( \vect{z}_2, P^{\left(Q_{1}, \lambda_{1}\right)}_{\vect{\Theta}| \vect{Z} = \vect{z}_{1}} \right), \qquad
\end{IEEEeqnarray} 
where the equality in~\eqref{EqFriendGoneToTexas1} follows from Lemma~\ref{LemmaSumLxy}; and 
the equality in~\eqref{EqFriendGoneToTexas2} follows from Definition~\ref{DefSensitivity}. The proof is completed by Theorem~\ref{TheoremSensitivityEqual}.
\end{IEEEproof}

\subsection{Averaging Gibbs Measures}

In practical scenarios, building GAs on the aggregated dataset might be difficult or impossible due to limited computational power or due to the fact that dataset aggregation at one location is not allowed due to privacy constraints. In these cases, a common practice is to average the output of machine learning algorithms trained on constituent datasets, e.g., federated learning \cite{mcmahan2017communication}. In this case, a figure of merit to validate such an approach is to study the difference of the expected empirical risk induced on the aggregated dataset by~$\mathsf{GA}_0$ and a convex combination of~$\mathsf{GA}_1$ and~$\mathsf{GA}_2$.  
The following theorem provides an explicit expression for this quantity. 

\begin{theorem}\label{TheoBlancNaiveStrong}
Assume that the~$\sigma$-finite measures~$Q_0$,~$Q_1$ and~$Q_2$ in~\eqref{EqERMRERi} are pair-wise mutually absolutely continuous.
Then, for all~$\alpha \in [0,1]$,
\begin{IEEEeqnarray}{rcl}
\nonumber
&& \mathsf{R}_{\vect{z}_{0}}\left(\alpha P^{\left(Q_{1}, \lambda_{1}\right)}_{\vect{\Theta}| \vect{Z} = \vect{z}_{1}} + \left( 1-\alpha\right)  P^{\left(Q_{2}, \lambda_{2}\right)}_{\vect{\Theta}| \vect{Z} = \vect{z}_{2}} \right)   - \mathsf{R}_{\vect{z}_{0}}\left( P^{\left(Q_{0}, \lambda_{0}\right)}_{\vect{\Theta}| \vect{Z} = \vect{z}_{0}} \right) \\
\nonumber
& = &\lambda_{0}\Bigg(  \KL{P^{\left(Q_{0}, \lambda_{0}\right)}_{\vect{\Theta}| \vect{Z} = \vect{z}_{0}}}{Q_{0}} \\
\nonumber
& & + \alpha \left( \KL{P^{\left(Q_1, \lambda_1\right)}_{\vect{\Theta}| \vect{Z} = \vect{z}_{1}}}{P^{\left(Q_{0}, \lambda_{0}\right)}_{\vect{\Theta}| \vect{Z} = \vect{z}_{0}}} - \KL{P^{\left(Q_{1}, \lambda_{1}\right)}_{\vect{\Theta}| \vect{Z} = \vect{z}_{1}}}{Q_{0}} \right)\\
\label{EqSheldonIsAmazing}
& & + \left( 1\hspace{-0.5ex} - \hspace{-0.5ex}\alpha\right) \left( \KL{P^{\left(Q_2, \lambda_2\right)}_{\vect{\Theta}| \vect{Z} = \vect{z}_{2}}}{P^{\left(Q_{0}, \lambda_{0}\right)}_{\vect{\Theta}| \vect{Z} = \vect{z}_{0}}} \hspace{-0.5ex}-\hspace{-0.5ex} \KL{P^{\left(Q_{2}, \lambda_{2}\right)}_{\vect{\Theta}| \vect{Z} = \vect{z}_{2}}}{Q_{0}} \right)\hspace{-1ex} \Bigg)\hspace{-.5ex}, \hspace{+3.5ex}
\end{IEEEeqnarray} 
where  the function~$\mathsf{R}_{\vect{z}_{0}}$ is defined in~\eqref{EqRxy} and, for all~$i \in \lbrace 1,2\rbrace$, the measure~$P^{\left(Q_{i}, \lambda_{i}\right)}_{\vect{\Theta}| \vect{Z} = \vect{z}_{i}}$ satisfies~\eqref{EqGenpdfi}.
\end{theorem}
\begin{IEEEproof}
The proof uses the following argument:
\begin{IEEEeqnarray}{rcl}
\nonumber
&& \mathsf{R}_{\vect{z}_{0}}\left(\alpha P^{\left(Q_{1}, \lambda_{1}\right)}_{\vect{\Theta}| \vect{Z} = \vect{z}_{1}} + \left( 1-\alpha\right)  P^{\left(Q_{2}, \lambda_{2}\right)}_{\vect{\Theta}| \vect{Z} = \vect{z}_{2}} \right)   - \mathsf{R}_{\vect{z}_{0}}\left( P^{\left(Q_{0}, \lambda_{0}\right)}_{\vect{\Theta}| \vect{Z} = \vect{z}_{0}} \right) \\
\nonumber
\nonumber
& = & \alpha \mathsf{R}_{\vect{z}_{0}}\left( P^{\left(Q_{1}, \lambda_{1}\right)}_{\vect{\Theta}| \vect{Z} = \vect{z}_{1}}  \right)   + \left( 1-\alpha\right)  \mathsf{R}_{\vect{z}_{0}}\left( P^{\left(Q_{2}, \lambda_{2}\right)}_{\vect{\Theta}| \vect{Z} = \vect{z}_{2}} \right) \\
\label{EqArepasFrescas1}
& &- \alpha \mathsf{R}_{\vect{z}_{0}}\left( P^{\left(Q_{0}, \lambda_{0}\right)}_{\vect{\Theta}| \vect{Z} = \vect{z}_{0}}  \right)   - \left( 1-\alpha\right)  \mathsf{R}_{\vect{z}_{0}}\left( P^{\left(Q_{0}, \lambda_{0}\right)}_{\vect{\Theta}| \vect{Z} = \vect{z}_{0}} \right)\\
\nonumber
& = & \alpha \left( \mathsf{R}_{\vect{z}_{0}}\left( P^{\left(Q_{1}, \lambda_{1}\right)}_{\vect{\Theta}| \vect{Z} = \vect{z}_{1}}  \right) - \mathsf{R}_{\vect{z}_{0}}\left( P^{\left(Q_{0}, \lambda_{0}\right)}_{\vect{\Theta}| \vect{Z} = \vect{z}_{0}}  \right) \right) \\
\label{EqArepasFrescas2}
& & + \left( 1-\alpha\right)  \left( \mathsf{R}_{\vect{z}_{0}}\left( P^{\left(Q_{2}, \lambda_{2}\right)}_{\vect{\Theta}| \vect{Z} = \vect{z}_{2}} \right) - \mathsf{R}_{\vect{z}_{0}}\left( P^{\left(Q_{0}, \lambda_{0}\right)}_{\vect{\Theta}| \vect{Z} = \vect{z}_{0}} \right) \right)\\
\label{EqArepasFrescas3}
& = & \alpha \mathsf{S}_{Q_0, \lambda_0}\left( \vect{z}_0, P^{\left(Q_{1}, \lambda_{1}\right)}_{\vect{\Theta}| \vect{Z} = \vect{z}_{1}} \right) \hspace{-0.5ex} + \hspace{-0.5ex}  \left( 1 \hspace{-0.5ex} - \hspace{-0.5ex} \alpha\right) \mathsf{S}_{Q_0, \lambda_0}\left( \vect{z}_0, P^{\left(Q_{2}, \lambda_{2}\right)}_{\vect{\Theta}| \vect{Z} = \vect{z}_{2}} \right), \qquad
\end{IEEEeqnarray} 
where the equality in~\eqref{EqArepasFrescas1} follows from Lemma~\ref{LemmaLinearity}, and 
the equality in~\eqref{EqArepasFrescas3} follows from Definition~\ref{DefSensitivity}. The proof is completed by Theorem~\ref{TheoremSensitivityEqual}.
\end{IEEEproof}

The following corollary of Theorem~\ref{TheoBlancNaiveStrong} is obtained by subtracting the equality in~\eqref{EqSheldonIsAmazing} with~$\alpha =1$ from the equality in~\eqref{EqSheldonIsAmazing} with~$\alpha =0$.

\begin{corollary}\label{CorBlancNaiveStrong}
Assume that the~$\sigma$-finite measures~$Q_0$,~$Q_1$ and~$Q_2$ in~\eqref{EqERMRERi} are pair-wise mutually absolutely continuous. Then, for all~$i \in \lbrace 0,1,2 \rbrace$, the probability measure~$P^{\left(Q_{i}, \lambda_{i}\right)}_{\vect{\Theta}| \vect{Z} = \vect{z}_{i}}$ in \eqref{EqGenpdfi} satisfies
\begin{IEEEeqnarray}{rcl}
\nonumber
&&\mathsf{R}_{\vect{z}_{0}}\left( P^{\left(Q_{2}, \lambda_{2}\right)}_{\vect{\Theta}| \vect{Z} = \vect{z}_{2}}\right) - \mathsf{R}_{\vect{z}_{0}}\left( P^{\left(Q_{1}, \lambda_{1}\right)}_{\vect{\Theta}| \vect{Z} = \vect{z}_{1}} \right)\nonumber\\ 
&  =  & \lambda_{0}  \Bigg(\KL{P^{\left(Q_2, \lambda_2\right)}_{\vect{\Theta}| \vect{Z} = \vect{z}_{2}}}{P^{\left(Q_{0}, \lambda_{0}\right)}_{\vect{\Theta}| \vect{Z} = \vect{z}_{0}}} - \KL{P^{\left(Q_{2}, \lambda_{2}\right)}_{\vect{\Theta}| \vect{Z} = \vect{z}_{2}} }{Q_0}\Bigg) \nonumber\\
\label{EqPizzaDeli6}
&    & - \lambda_{0} \Bigg( \KL{P^{\left(Q_1, \lambda_1\right)}_{\vect{\Theta}| \vect{Z} = \vect{z}_{1}}}{P^{\left(Q_{0}, \lambda_{0}\right)}_{\vect{\Theta}| \vect{Z} = \vect{z}_{0}}}   -  \KL{P^{\left(Q_{1}, \lambda_{1}\right)}_{\vect{\Theta}| \vect{Z} = \vect{z}_{1}}}{Q_{0}} \Bigg), \IEEEeqnarraynumspace
\end{IEEEeqnarray}
where  the function~$\mathsf{R}_{\vect{z}_{0}}$ is defined in~\eqref{EqRxy}.
\end{corollary}

Corollary~\ref{CorBlancNaiveStrong} is an alternative to Theorem~\ref{TheoThreeYearAnniversary} involving the GA trained upon the aggregated dataset, i.e.,~$\mathsf{GA}_0$.

\subsection{Gibbs Algorithms Trained on Aggregated Datasets}
Training a GA upon the aggregation of datasets does not necessarily imply lower expected empirical risk on the constituent datasets. As argued before, datasets might be obtained up to different levels of fidelity. Hence, a validation  
method for~$\mathsf{GA}_0$ is based on the expected empirical risk induced by~$\mathsf{GA}_0$ on a constituent dataset~$\vect{z}_{i}$, with~$i \in \lbrace 1,2 \rbrace$,
which is denoted by~$\mathsf{R}_{\vect{z}_{i}}\left( P^{\left(Q_{0}, \lambda_{0}\right)}_{\vect{\Theta}| \vect{Z} = \vect{z}_{0}} \right)$. A pertinent figure of merit is the difference~$\mathsf{R}_{\vect{z}_{i}}\left( P^{\left(Q_{0}, \lambda_{0}\right)}_{\vect{\Theta}| \vect{Z} = \vect{z}_{0}} \right) - \mathsf{R}_{\vect{z}_{i}}\left( P^{\left(Q_{i}, \lambda_{i}\right)}_{\vect{\Theta}| \vect{Z} = \vect{z}_{i}} \right)$. The following theorem provides an explicit expression for such quantity. 
\begin{theorem}\label{TheoTallNaiveStrong}
Assume that the~$\sigma$-finite measures~$Q_0$,~$Q_1$ and~$Q_2$ in~\eqref{EqERMRERi} are pair-wise mutually absolutely continuous. Then, for all~$i \in \lbrace 0,1,2 \rbrace$, 
\begin{IEEEeqnarray}{rcl}
\nonumber
& &\mathsf{R}_{\vect{z}_{i}} \left( P^{\left(Q_{0}, \lambda_{0}\right)}_{\vect{\Theta}| \vect{Z} = \vect{z}_{0}} \right) -  \mathsf{R}_{\vect{z}_{i}}\left( P^{\left(Q_{i}, \lambda_{i}\right)}_{\vect{\Theta}| \vect{Z} = \vect{z}_{i}} \right)  =  \lambda_{i} \bigg( \KL{P^{\left(Q_{i}, \lambda_{i}\right)}_{\vect{\Theta}| \vect{Z} = \vect{z}_{i}}}{Q_{i}} \\
\label{EqNapoli0876}
&  & + \KL{P^{\left(Q_{0}, \lambda_{0}\right)}_{\vect{\Theta}| \vect{Z} = \vect{z}_{0}}}{P^{\left(Q_i, \lambda_i\right)}_{\vect{\Theta}| \vect{Z} = \vect{z}_{i}}}  - \KL{P^{\left(Q_{0}, \lambda_{0}\right)}_{\vect{\Theta}| \vect{Z} = \vect{z}_{0}}}{Q_{i}} \bigg) ,\quad
\end{IEEEeqnarray}
where,  the function~$\mathsf{R}_{\vect{z}_{i}}$ is defined in~\eqref{EqRxy} and the measure~$P^{\left(Q_{i}, \lambda_{i}\right)}_{\vect{\Theta}| \vect{Z} = \vect{z}_{i}}$ satisfies~\eqref{EqGenpdfi}.
\end{theorem}
\begin{IEEEproof}
The proof is immediate from Theorem~\ref{TheoremSensitivityEqual} by noticing that for all~$i \in \lbrace 1,2 \rbrace$, the differences~$\mathsf{R}_{\vect{z}_{i}}\left( P^{\left(Q_{0}, \lambda_{0}\right)}_{\vect{\Theta}| \vect{Z} = \vect{z}_{0}}\right) - \mathsf{R}_{\vect{z}_{i}}\left( P^{\left(Q_{i}, \lambda_{i}\right)}_{\vect{\Theta}| \vect{Z} = \vect{z}_{i}}\right)$ can be written in terms of the sensitivity~$\mathsf{S}_{Q_i, \lambda_i}\left( \vect{z}_i, P^{\left(Q_{0}, \lambda_{0}\right)}_{\vect{\Theta}| \vect{Z} = \vect{z}_{0}} \right)$.
\end{IEEEproof} 

\subsection{Special Cases}\label{SecSpecialCases}
Consider a given~$\sigma$-finite measure~$Q$ and assume that for all~$i \in \lbrace 0, 1, 2 \rbrace$ and for all~$\set{A} \in \BorSigma{\set{M}}$,~$Q\left( \set{A} \right)  = Q_i\left( \set{A} \right)$. Assume also that the parameters~$\lambda_0$,~$\lambda_1$,  and~$\lambda_2$ in~\eqref{EqERMRERi} satisfy  
$\lambda_1 =  \frac{n_{0}}{n_1}\lambda_0$ and~$\lambda_2 =  \frac{n_{0}}{n_2}\lambda_0$. 
%
These assumptions are referred to as the case of \emph{homogeneous priors} with measure~$Q$, and the case of \emph{proportional regularization}, respectively. The term ``proportional'' stems from the fact that the regularization factor decreases proportionally to the size of the data set in the optimization problem in~\eqref{EqERMRERi}.
Under these assumptions, the following corollary of Theorem~\ref{TheoBlancNaive} unveils an interesting connection with the Jeffrey's divergence~\cite{jeffreys1946invariant}.
\begin{corollary}\label{CorBlancNaiveTop}
Consider the case of homogeneous priors with a $\sigma$-finite measure $Q$ and proportional regularization with parameter $\lambda_0$. Then, for all $i \in \lbrace 1,2 \rbrace$, the probability measure $P^{\left(Q, \lambda_{i}\right)}_{\vect{\Theta}| \vect{Z} = \vect{z}_{i}}$ in~\eqref{EqGenpdfi}, satisfies
\begin{IEEEeqnarray}{l}
\nonumber
  \Big( \frac{n_1}{n_0} \mathsf{R}_{\vect{z}_{1}}\left( P^{\left(Q, \lambda_{2}\right)}_{\vect{\Theta}| \vect{Z} = \vect{z}_{2}}\right) -  \frac{n_2}{n_0} \mathsf{R}_{\vect{z}_{2}}\left( P^{\left(Q, \lambda_{2}\right)}_{\vect{\Theta}| \vect{Z} = \vect{z}_{2}} \right)  \Big) \\ 
\nonumber
+ \Big(  \frac{n_2}{n_0}  \mathsf{R}_{\vect{z}_{2}}\left( P^{\left(Q, \lambda_{1}\right)}_{\vect{\Theta}| \vect{Z} = \vect{z}_{1}} \right) - \frac{n_1}{n_0} \mathsf{R}_{\vect{z}_{1}}\left( P^{\left(Q, \lambda_{1}\right)}_{\vect{\Theta}| \vect{Z} = \vect{z}_{1}} \right)  \Big)  \\
\label{EqPizzaHomog0876}
 =   \lambda_{0} \Bigg( \KL{P^{\left(Q, \lambda_1\right)}_{\vect{\Theta}| \vect{Z} = \vect{z}_{1}}}{P^{\left(Q, \lambda_2\right)}_{\vect{\Theta}| \vect{Z} = \vect{z}_{2}}}  + \KL{P^{\left(Q, \lambda_2\right)}_{\vect{\Theta}| \vect{Z} = \vect{z}_{2}}}{P^{\left(Q, \lambda_1\right)}_{\vect{\Theta}| \vect{Z} = \vect{z}_{1}}} \Bigg). \middlesqueezeequ \IEEEeqnarraynumspace
\end{IEEEeqnarray}
\end{corollary}
Note that $\KL{P^{\left(Q, \lambda_1\right)}_{\vect{\Theta}| \vect{Z} = \vect{z}_{1}}}{P^{\left(Q, \lambda_2\right)}_{\vect{\Theta}| \vect{Z} = \vect{z}_{2}}}  + \KL{P^{\left(Q, \lambda_2\right)}_{\vect{\Theta}| \vect{Z} = \vect{z}_{2}}}{P^{\left(Q, \lambda_1\right)}_{\vect{\Theta}| \vect{Z} = \vect{z}_{1}}}$ is the Jeffrey's divergence between the measures $P^{\left(Q, \lambda_1\right)}_{\vect{\Theta}| \vect{Z} = \vect{z}_{1}}$ and $P^{\left(Q, \lambda_2\right)}_{\vect{\Theta}| \vect{Z} = \vect{z}_{2}}$.
For all $i \in \lbrace 1,2 \rbrace$ and $j \in \lbrace 1,2 \rbrace\setminus\lbrace i \rbrace$,  the difference 
$\frac{n_j}{n_0} \mathsf{R}_{\vect{z}_{j}}\left( P^{\left(Q, \lambda_{i}\right)}_{\vect{\Theta}| \vect{Z} = \vect{z}_{i}}\right) -  \frac{n_i}{n_0} \mathsf{R}_{\vect{z}_{i}}\left( P^{\left(Q, \lambda_{i}\right)}_{\vect{\Theta}| \vect{Z} = \vect{z}_{i}} \right)$  
is reminiscent of a \emph{validation} \cite[Section $11.2$]{shalev2014understanding}. This follows from noticing that $\mathsf{R}_{\vect{z}_{j}}\left( P^{\left(Q, \lambda_{i}\right)}_{\vect{\Theta}| \vect{Z} = \vect{z}_{i}}\right)$ is the testing error of~$\mathsf{GA}_i$ over the test dataset~$\vect{z}_{j}$, while $ \mathsf{R}_{\vect{z}_{i}}\left( P^{\left(Q, \lambda_{i}\right)}_{\vect{\Theta}| \vect{Z} = \vect{z}_{i}} \right)$ is the training error of~$\mathsf{GA}_i$.

In \eqref{EqPizzaHomog0876}, it holds that $\KL{P^{\left(Q, \lambda_1\right)}_{\vect{\Theta}| \vect{Z} = \vect{z}_{1}}}{P^{\left(Q, \lambda_2\right)}_{\vect{\Theta}| \vect{Z} = \vect{z}_{2}}}$ and $\KL{P^{\left(Q, \lambda_2\right)}_{\vect{\Theta}| \vect{Z} = \vect{z}_{2}}}{P^{\left(Q, \lambda_1\right)}_{\vect{\Theta}| \vect{Z} = \vect{z}_{1}}}$ are both nonnegative, which leads to the following corollary of Theorem~\ref{TheoBlancNaive}.
\begin{corollary}\label{CorBlancNaiveHomog}
Consider the case of homogeneous priors with a $\sigma$-finite measure $Q$ and proportional regularization. Then, for all $i \in \lbrace 1,2 \rbrace$, the probability measure $P^{\left(Q, \lambda_{i}\right)}_{\vect{\Theta}| \vect{Z} = \vect{z}_{i}}$ in~\eqref{EqGenpdfi}, satisfies
\begin{IEEEeqnarray}{rcl}
\nonumber
& & \Big( \frac{n_1}{n_0} \mathsf{R}_{\vect{z}_{1}}\left( P^{\left(Q, \lambda_{2}\right)}_{\vect{\Theta}| \vect{Z} = \vect{z}_{2}}\right) +  \frac{n_2}{n_0} \mathsf{R}_{\vect{z}_{2}}\left( P^{\left(Q, \lambda_{1}\right)}_{\vect{\Theta}| \vect{Z} = \vect{z}_{1}} \right)  \Big)  \\
&  \geqslant  & \Big( \frac{n_1}{n_0} \mathsf{R}_{\vect{z}_{1}}\left( P^{\left(Q, \lambda_{1}\right)}_{\vect{\Theta}| \vect{Z} = \vect{z}_{1}} \right) +  \frac{n_2}{n_0}  \mathsf{R}_{\vect{z}_{2}}\left( P^{\left(Q, \lambda_{2}\right)}_{\vect{\Theta}| \vect{Z} = \vect{z}_{2}} \right) \Big).
\end{IEEEeqnarray}
\end{corollary}
Corollary~\ref{CorBlancNaiveHomog} highlights that the weighted-sum of the test errors induced by~$\mathsf{GA}_1$ and~$\mathsf{GA}_2$ is not smaller than the weighted sum of their training errors when the weights are proportional to the sizes of the datasets. 
 
\IEEEtriggeratref{18}
\bibliographystyle{IEEEtran}
\bibliography{references}
 \end{document}